\documentclass{article}

\PassOptionsToPackage{authoryear, round, sort}{natbib}

\usepackage[dblblindworkshop, final]{neurips_2025}
\workshoptitle{Machine Learning and the Physical Sciences}

\usepackage[utf8]{inputenc} %
\usepackage[T1]{fontenc}    %
\usepackage{hyperref}       %
\usepackage{url}            %
\usepackage{booktabs}       %
\usepackage{amsfonts}       %
\usepackage{nicefrac}       %
\usepackage{microtype}      %
\usepackage{xcolor}         %
\usepackage{amsmath} 
\usepackage{bbm}   %
\title{\emph{JaxWildfire} -- A GPU-Accelerated Wildfire Simulator for Reinforcement Learning }

\usepackage{subcaption} %
\usepackage{xcolor}

\usepackage{wrapfig}
\definecolor{wcTree}{HTML}{006400}
\definecolor{wcShrub}{HTML}{FFBB22}
\definecolor{wcGrass}{HTML}{FFFF4C}
\definecolor{wcCrop}{HTML}{F096FF}
\definecolor{wcBuilt}{HTML}{800080}
\definecolor{wcBare}{HTML}{B4B4B4}
\definecolor{wcSnow}{HTML}{F0F0F0}
\definecolor{wcWater}{HTML}{0064C8}
\definecolor{wcWetland}{HTML}{0096A0}
\definecolor{wcMangrove}{HTML}{00CF75}
\definecolor{wcLichen}{HTML}{FAE6A0}

\author{%
  Ufuk Çakır \\
  \And
  Victor-Alexandru Darvariu\\
  \And
  Bruno Lacerda\\
  \And
  Nick Hawes \\ 
  \\
  Oxford Robotics Institute\\
  University of Oxford\\
  United Kingdom \\
  \texttt{\{ufukcakir,victord,bruno,nickh\}@robots.ox.ac.uk} \\
}

\newcommand{\jwf}{\emph{JaxWildfire}}

\usepackage{layouts}
\usepackage{graphicx}

\begin{document}

\maketitle

\begin{abstract}
Artificial intelligence methods are increasingly being explored for managing wildfires and other natural hazards. 
In particular, reinforcement learning (RL) is a promising path towards improving outcomes in such uncertain decision-making scenarios and moving beyond reactive strategies.
However, training RL agents requires many environment interactions, and the speed of existing wildfire simulators is a severely limiting factor. We introduce \jwf{}, a simulator underpinned by a principled probabilistic fire spread model based on cellular automata. It is implemented in JAX and enables vectorized simulations using \texttt{vmap}, allowing high throughput of simulations on GPUs. We demonstrate that \jwf{} achieves 6-35x speedup over existing software and enables gradient-based optimization of simulator parameters. 
Furthermore, we show that \jwf{} can be used to train RL agents to learn wildfire suppression policies. Our work is an important step towards enabling the advancement of RL techniques for managing natural hazards. 

\end{abstract}
\section{Motivation}
\begin{wrapfigure}{r}{0.42\textwidth}
  \centering
  \vspace{-4.1\baselineskip}
  \includegraphics[width=0.42\textwidth]{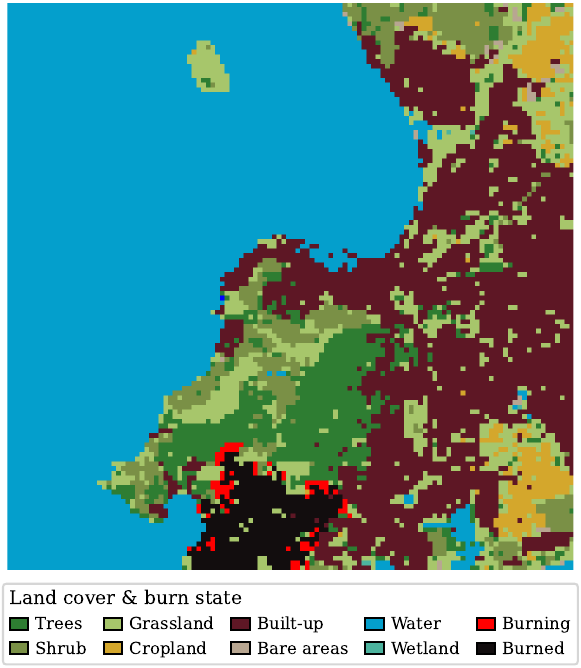}
  \caption{\jwf{} simulation in Cape Town, SA using ESA WorldCover data.}
  \label{fig:logo}
  \vspace{-5mm}
\end{wrapfigure}

While AI has found important uses for natural hazard management such as damage assessment and forecasting, experts call for a fundamental paradigm shift from reactive pattern recognition to anticipatory sequential decision-making under uncertainty \citep{2020a,beduschi2022}. Reinforcement learning (RL) methods offer a promising framework for such settings; however, they require millions of environment interactions to learn effective policies. This makes high-fidelity wildfire simulators like FARSITE \citep{finney1998} and ELMFIRE \citep{lautenberger2013} too expensive for RL applications.

Even though existing cellular automata (CA) based simulators are faster, they are still fundamentally limited by the requirement to transfer data to main memory in the interaction between the simulation and RL agent components. Recent tools like JAX \citep{jax2018github}, which allow performing the environment simulation and the RL algorithm training jointly on GPUs, are quickly becoming the de facto standard for RL research, as they allow for efficient memory management and faster training times.

In this paper, we introduce \jwf{}, which provides similar functionality to existing CA simulators while being an order of magnitude faster. We demonstrate that gradients can flow through the simulator, enabling gradient-based optimization of simulator parameters based on historical wildfire data. \jwf{} is also compatible with the European Space Agency's WorldCover dataset \citep{zanaga_2021_5571936} and Digital Elevation Model (DEM) \citep{ESA_CopernicusDEM_GLO30}, allowing the creation of realistic environments by simply providing bounding box coordinates or names of cities.

\jwf{} is compatible with the JAX-based RL library Gymnax \citep{gymnax2022github}, allowing convenient integration with existing RL algorithm implementations. We successfully validate the suitability of \jwf{} for training RL agents by studying a synthetic scenario in which an air tanker obtains rewards for extinguishing a developing fire. Our work contributes a societally important simulation scenario for the RL community to work on and serves as an important enabling step towards advancing RL-based methodologies for natural hazard management.

\section{Related Work}
Wildfire simulators usually fall into two categories: physics-based models used in operations and research and cellular automata (CA) aimed at speed and scale \citep{papadopoulos2011}.

\textbf{Physics-based models.}
FARSITE \citep{finney1998} and ELMFIRE \citep{lautenberger2013} are two widely used physics-based wildfire simulators that target high fidelity. Although they are employed operationally, their computational costs limit their use in RL research.

\textbf{Cellular automata.} CA models approximate fire spread using simple, local rules, making them computationally efficient and suitable for large-scale simulations.
They have been widely adopted in wildfire research \citep{rui2018, pais2019,karafyllidis1997}.
The most recent work is PyTorchFire \citep{xia2025}, a GPU accelerated simulator which allows for parameter calibration with PyTorch \citep{pytorch}, and is based on the CA model of \citet{alexandridis2008a}. They demonstrate that their model can be calibrated using gradient-based optimization and achieves promising results on real wildfire data \citep{xia2024data}.

\textbf{RL-oriented interfaces.} Recent works have emerged that specifically target the RL community.
SimFire \citep{Doyle_SimFire_2024} simulates wildfire spread using Rothermel equations \citep{rothermel1972} and uses environment data from LANDFIRE \citep{landfire2021models}. SimHarness \citep{tapley2023} wraps SimFire and provides a convenient interface for RL research to train agents. However, implementation speed is a limiting factor for the scale (and, consequently, task complexity) of training RL agents. Due to this, RL research is increasingly shifting towards JAX as the main framework \citep{rutherford2024a}. To the best of our knowledge, \jwf{} is the first wildfire simulator implemented in JAX that is specifically designed for RL research.

\section{\jwf{} Simulator}
\label{sec:jwf}
\begin{figure}
    \centering
    \includegraphics[width=\textwidth]{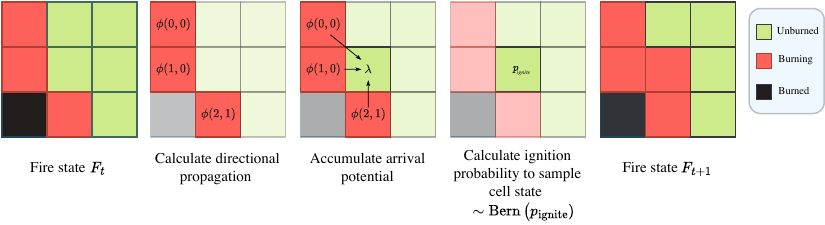}
    \caption{\textit{Fire spread visualization}. Our fire spread mechanism based on cellular automata propagates fire directionally. The arrival potential is accumulated from neighboring cells and used to determine the ignition probability. The state of the cell at the next timestep is sampled from a Bernoulli distribution parameterized by the ignition probability.}
    \label{fig:placeholder}
\end{figure}
The core of \jwf{} is a CA model that simulates fire spread as a function of wind, slope, vegetation, and fuel density.
Our model builds on \citet{alexandridis2008a} and \citet{xia2025}, but introduces a more realistic model of ignition probability and a principled conversion from fire potential to ignition probability.\\
\textbf{State representation.}
At timestep $t$ the simulator holds four layers with shape $(H,W)$ encoding information about each cell \( u = (i,j) \).
\begin{itemize}
  \item \textbf{Fire state} -- $F_t(u)\in\{0,1,2\}$  
        ($0=$\,unburned,\; $1=$\,burning,\; $2=$\,burned).
  \item \textbf{Wind} --  
        speed $V_t(u)\in\mathbb{R}_{\ge0}$ and direction
        $\Theta_t(u)\in[0,2\pi)$.
  \item \textbf{Vegetation} --
Each cell carries vegetation–related fields
\(
\mu_{\text{veg}}(u)\;,
\mu_{\text{den}}(u)\;
\) where
\begin{itemize}
  \item \(\mu_{\text{veg}}(u)\in[-1,1]\) – canopy ignition modifier, where positive values favor burning.
      \item \(\mu_{\text{den}}(u)\in[-1,1]\) – ground-fuel density modifier, ranges from $-1$ (no fuel) to $1$ (dense).
\end{itemize}
  \item \textbf{Slope} --
    $S(u)\in\mathbb{R}^{3\times 3}$ stores the slope of the terrain in the direction of each neighbor.
      Positive values indicate uphill in the propagation direction,
      negative values downhill.
\end{itemize}
The slope is assumed to be static as the elevation in the environment does not change, but the remaining fields are time-dependent.\\
\textbf{Directional propagation.}
For a burning source cell \(u=(i,j)\), the potential for fire to propagate to a neighboring cell
is modeled by an unbounded dimensionless score \( \phi\) as a function of the wind, slope, vegetation, and fuel density.
The influence of wind and slope are modeled as directional propagation factors in the direction of the neighbor \(\Delta = (\delta x ,\delta y)\in\{-1,0,1\}^2\) with
\begin{align}
  \kappa_{wind}(u, \Delta) &= \exp(\alpha_{w1} V(u)) \cdot \exp\bigl(\alpha_{w2} V(u) [\cos(\varphi(\Delta) - \Theta(u)) - 1]\bigr) \\
  \kappa_{slope}(u, \Delta) &= \exp(\alpha_{s} S(u, \Delta))
\end{align}
where \(\varphi(\Delta)\) is the angle (where east equals zero, counting counter-clockwise) of the offset \(\Delta\)  and \(\alpha_{w1}, \alpha_{w2}, \alpha_s\) are tunable parameters that 
control the influence of wind and slope on fire spread. The total propagation potential from source cell \(u\) in direction \(\Delta\) is then given by 
\begin{equation}
  \phi(u, \Delta) = p_\text{base} \cdot (1 + \mu_{\text{veg}}(u)) \cdot (1 + \mu_{\text{den}}(u)) \cdot \kappa_{wind}(u, \Delta) \cdot \kappa_{slope}(u, \Delta)
  \label{eq:propagation}
\end{equation}
where \(p_\text{base}\) is a base propagation factor that can be tuned to control the overall fire spread rate.
This follows the model of \citet{xia2025}.

\textbf{Ignition probability.}
We model the ignition of a cell as a two-step process: first, we calculate the probability of fire arriving at a target cell \( v \) from its neighbors, and then we calculate the probability of ignition based on the target cell's fuel properties.
Rather than using an ad-hoc squashing function \( \tanh \) like \cite{xia2025}, we interpret
the propagation potential \( \phi(u, \Delta)\) as the intensity rate \( \lambda \) of a Poisson process.
The intensity rate is the expected number of fire sparks arriving at a target cell \( v \) from its neighbors \( u\in\mathcal{N}(v)\) in a time step.
Multiplying by the susceptibility \( \gamma(v) \) of the target cell \( v \) gives the effective Poisson rate of successful ignition events.
The susceptibility is a function of the target cell's vegetation and fuel density \( \gamma(v) = \alpha_\gamma(1 + \mu_{\text{veg}}(v)) \cdot (1 + \mu_{\text{den}}(v))\).
The probability of a target cell \( v \) igniting is therefore
\begin{equation}
  p_\text{ignite}(v) = 1 - \exp\bigl(-\gamma(v) \lambda\bigr) \qquad \text{with} \quad \lambda = \sum_{u\in\mathcal{N}(v)} b(u)\phi(u, \Delta_{u \rightarrow v})
  \label{eq:propagation2}
\end{equation}
where \( b(u) \) is the burning state of the neighbor cell \( u \) (1 if burning, 0 otherwise) and \(\Delta_{u \rightarrow v}\) is the direction from \( u \) to \( v\).
This provides a differentiable conversion that ensures probability is bounded in \([0,1]\) and can be used in gradient-based optimization of the wildfire simulator parameters.

\textbf{Stochastic update} $F_t\!\rightarrow\!F_{t+1}$.
Unburned cells ignite with probability $p_{\text{ignite}}(v)$.  
Burning cells remain burning with probability $p_{\text{continue}}$; otherwise they transition to \textit{Burned}.  
The simulator has two routines: it can propagate the raw probabilities (deterministic and continuous) or draw independent Bernoulli samples for these events (stochastic and discrete).

\paragraph{Deterministic (raw-probabilities) update.}
Let the per-cell continuous state at time \(t\) be
\[
\mathbf{p}_t(v) = \bigl(p^{\text{un}}_t(v),\; p^{\text{burn}}_t(v),\; p^{\text{bd}}_t(v)\bigr),
\qquad
p^{\text{un}}_t + p^{\text{burn}}_t + p^{\text{bd}}_t = 1,
\]
where \(p^{\text{un}}\) is the probability the cell is unburned, \(p^{\text{burn}}\) the probability it is burning, and \(p^{\text{bd}}\) the probability it is burned-out.  
Given the ignition probability \(p_{\mathrm{ignite}}(v)\) from Eq.~\eqref{eq:propagation2} and the continuation probability \(p_{\mathrm{continue}}\), the one-step update is written elementwise as
\begin{align}
\text{newly\_burning}_t(v) &= p^{\text{un}}_t(v)\; p_{\mathrm{ignite}}(v), \\
\text{continuing}_t(v)    &= p^{\text{burn}}_t(v)\; p_{\mathrm{continue}}, \\
\text{burned\_now}_t(v)   &= p^{\text{burn}}_t(v)\; \bigl(1 - p_{\mathrm{continue}}\bigr),
\end{align}
and the state evolves by
\begin{align}
p^{\text{burn}}_{t+1}(v) &= \text{newly\_burning}_t(v) \;+\; \text{continuing}_t(v),\label{eq:det_burn}\\
p^{\text{un}}_{t+1}(v)   &= p^{\text{un}}_t(v) \;-\; \text{newly\_burning}_t(v),\label{eq:det_un}\\
p^{\text{bd}}_{t+1}(v)   &= p^{\text{bd}}_t(v) \;+\; \text{burned\_now}_t(v).\label{eq:det_burned}
\end{align}

\textbf{Data integrations.} \jwf{} supports synthetic randomly generated forest environments as well as realistic environments based on real-world data from the European Space Agency's WorldCover dataset \citep{zanaga_2021_5571936} and Digital Elevation Model (DEM) data \citep{ESA_CopernicusDEM_GLO30}. 
The user only needs to provide a place-name string (e.g "Cape Town, South Africa") which is then internally geocoded into a bounding box and used to download the WorldCover and DEM tiles.
Weather data such as wind speed and direction can be provided as static input or as time series data to simulate changing weather conditions. Lastly, \jwf{} also provides functionality to export data to PyTorch format, allowing the benchmarking of machine learning models for learning fire spread dynamics.
\begin{figure}[t]
  \centering
  \includegraphics[width=\textwidth]{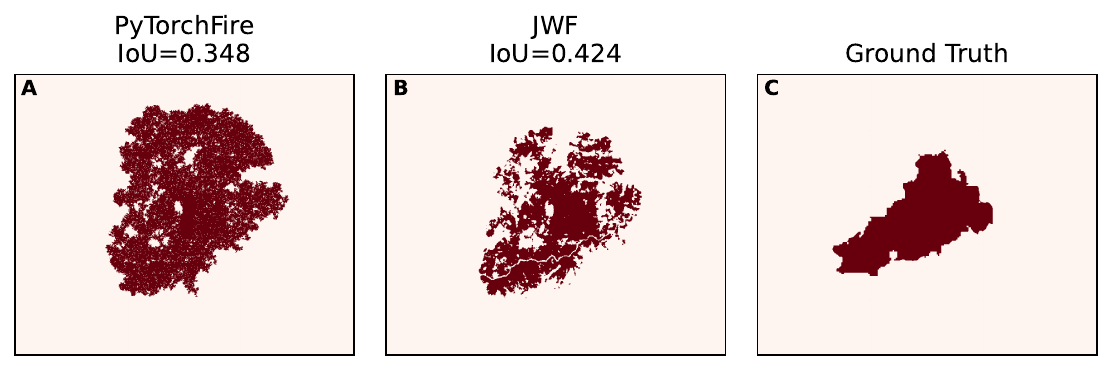}
  \caption{\textit{Calibration on a real wildfire.} Comparison of predicted fire spread of Bear 2020 wildfire \citep{xia2024data} averaged over 5 seeds (variance was negligible). The panels show the spread of the fire as simulated by \textbf{A)} PyTorchFire after calibration; \textbf{B)} \jwf{} after calibration; \textbf{C)} real historical fire perimeter. The simulators achieve comparable results measured by Intersection over Union, with a slight advantage for \jwf{}.}
  \label{fig:real_calibration}
\end{figure}

\section{Results}
\begin{figure}[h!]
  \centering
  \begin{subfigure}[t]{0.48\textwidth}
    \centering
    \includegraphics[width=\textwidth]{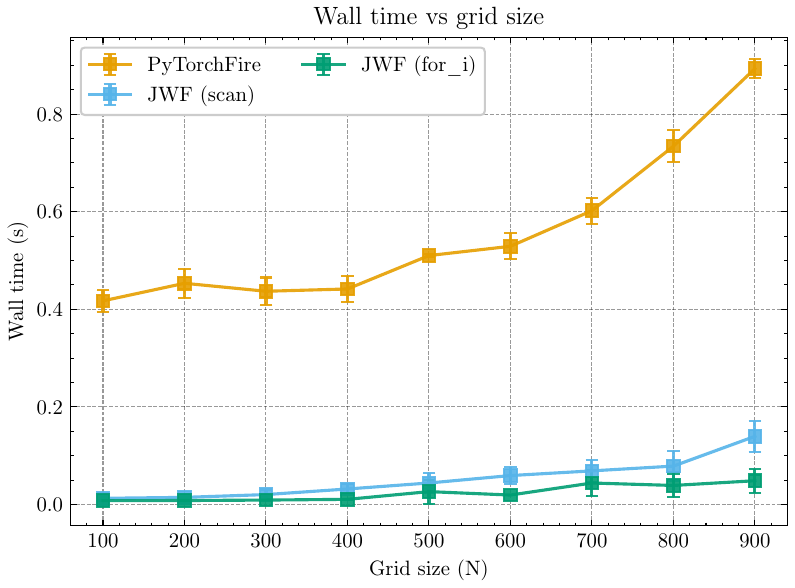}
    \label{fig:walltime}
  \end{subfigure}
  \hfill
  \begin{subfigure}[t]{0.48\textwidth}
    \centering
    \includegraphics[width=\textwidth]{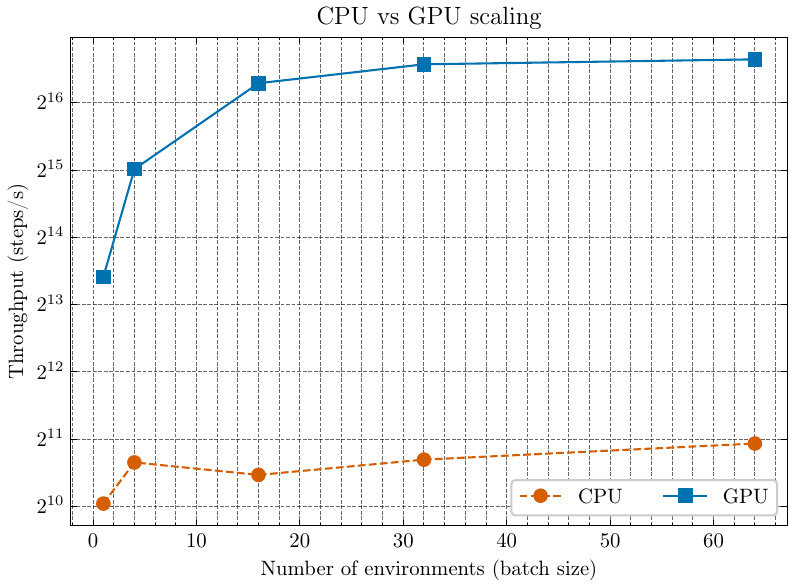}
    \label{fig:scaling_gpu}
  \end{subfigure}
  \caption{Performance evaluation of \jwf{}. Left: wall time (s) on GPU, showing \jwf{} is an order of magnitude faster than PyTorchFire, the competing library. Right: throughput of \jwf{} on GPU versus CPU, demonstrating that our framework benefits from increasing degrees of parallelism.}
  \label{fig:scaling_comparison}
\end{figure}

We use PyTorchFire as baseline as it is the closest related work to \jwf{}.
We benchmark both simulators on an NVIDIA A100 GPU on different grid sizes and 5 different random seeds. 
We report the mean and standard deviation of the time taken to simulate 200 time steps in Figure~\ref{fig:scaling_comparison}.
\jwf{} achieves a speedup of $6-35\times$ over PyTorchFire for single simulations, which goes up to $18-51\times$ when using \texttt{for\_i}, which does not track history and is therefore more efficient. 
Additionally, \jwf{} can leverage JAX's \texttt{vmap} to run multiple vectorized simulations on the GPU, achieving a peak throughput of \( 2.5^{16}\) steps/s with $64$ environments on a single A100 GPU and a grid size of \(64\times 64\) (see Figure \ref{fig:scaling_comparison}).

To show that the simulator can be calibrated using gradient-based optimization, we follow the setup of \citet{xia2025} and use a loss function that combines binary cross-entropy (BCE) loss to measure the difference in fire spread shape and a mean squared error (MSE) loss after 2D average pooling to measure the difference in fire scale.
We use wildfire data from Bear 2020 \citep{xia2024data} to calibrate both simulators and report the results of the best runs in Figure~\ref{fig:real_calibration}.
For \jwf{} we use the \texttt{optax} \citep{deepmind2020jax} library to perform gradient-based optimization using Adam \citep{kingma2017} with the raw probabilities routine. \jwf{} achieves marginally better results in simulation accuracy compared to PyTorchFire.

\begin{wrapfigure}{l}{0.5\textwidth}
  \vspace{0.5\baselineskip}
  \centering
  \begin{center}
    \includegraphics[width=0.49\textwidth]{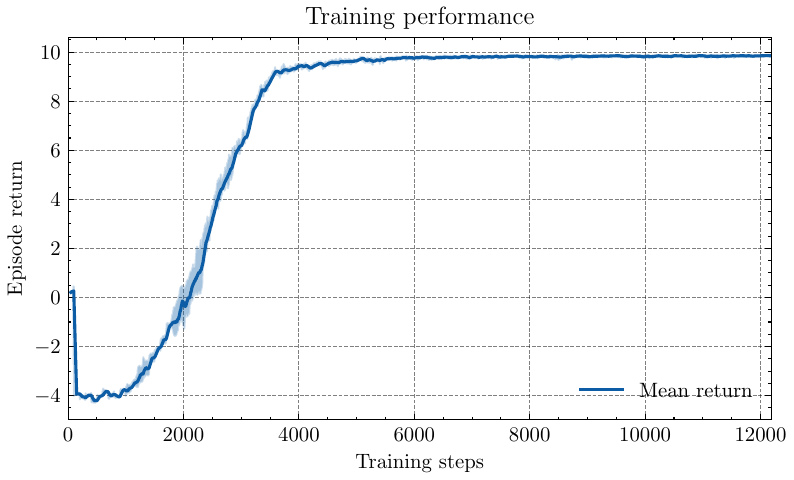}
  \end{center}
  \caption{PPO training curve in a synthetic 20x20 environment. A reward of 10 is given for completely extinguishing the fire.}
  \vspace{-1\baselineskip}
  \label{fig:ppo_training_curve}
\end{wrapfigure}

To demonstrate the suitability of \jwf{} for RL research, we trained an air tanker agent in a synthetic 20x20 environment using  PPO \citep{schulman2017,lu2022discovered} to learn a fire suppression policy. 
The agent receives partial egocentric observations and controls the valve (open/closed) to drop limited water resources. Water transforms the state of a cell from burning to burned. 
We give a penalty at each timestep proportional to the number of burning cells and a terminal \( +10 \) reward if the fire is completely extinguished.
During training, we randomize initial fire and agent positions and use fixed wind conditions.
The training curve in Figure~\ref{fig:ppo_training_curve} shows that the agent learns to extinguish the fire in this scenario. Visualizations of the trained agent acting in the environment are provided in Figure~\ref{fig:env_illustration} in the Appendix.

\section{Conclusion, Limitations and Outlook }
In this work, we have proposed \jwf{}, a fast and efficient JAX-native wildfire simulator for RL research that achieves a significant speedup of $6-35\times$ over a recent comparable framework. It enables differentiable tuning of simulator parameters and is integrated with real-world data sources. We have demonstrated its suitability for training RL agents on a synthetic fire extinction scenario. Future work includes adding support for heterogeneous multi-agent teams, such as ground vehicles and drones, to simulate more complex and practical wildfire management situations. We also aim to extend the underlying model to consider the effects of humidity and temperature on fire spread. Our work contributes a societally relevant benchmark problem for the RL community and also to broader efforts to develop proactive decision support systems for natural hazard management. 

\newpage
\bibliographystyle{plainnat}
\bibliography{references}
\clearpage
\section*{Appendix}

\textbf{Code availability.} We aim to publicly release \jwf{} with a permissive license such that it can be used by the community. Our repository follows good software engineering practices, featuring documentation and unit tests. The repository can be found under \url{https://github.com/ori-goals/JaxWildfire}.

\textbf{Additional figures.}  Additional figures, referenced in the main text, can be found below.

\begin{figure}[h!]
  \centering
  \includegraphics[width=\textwidth]{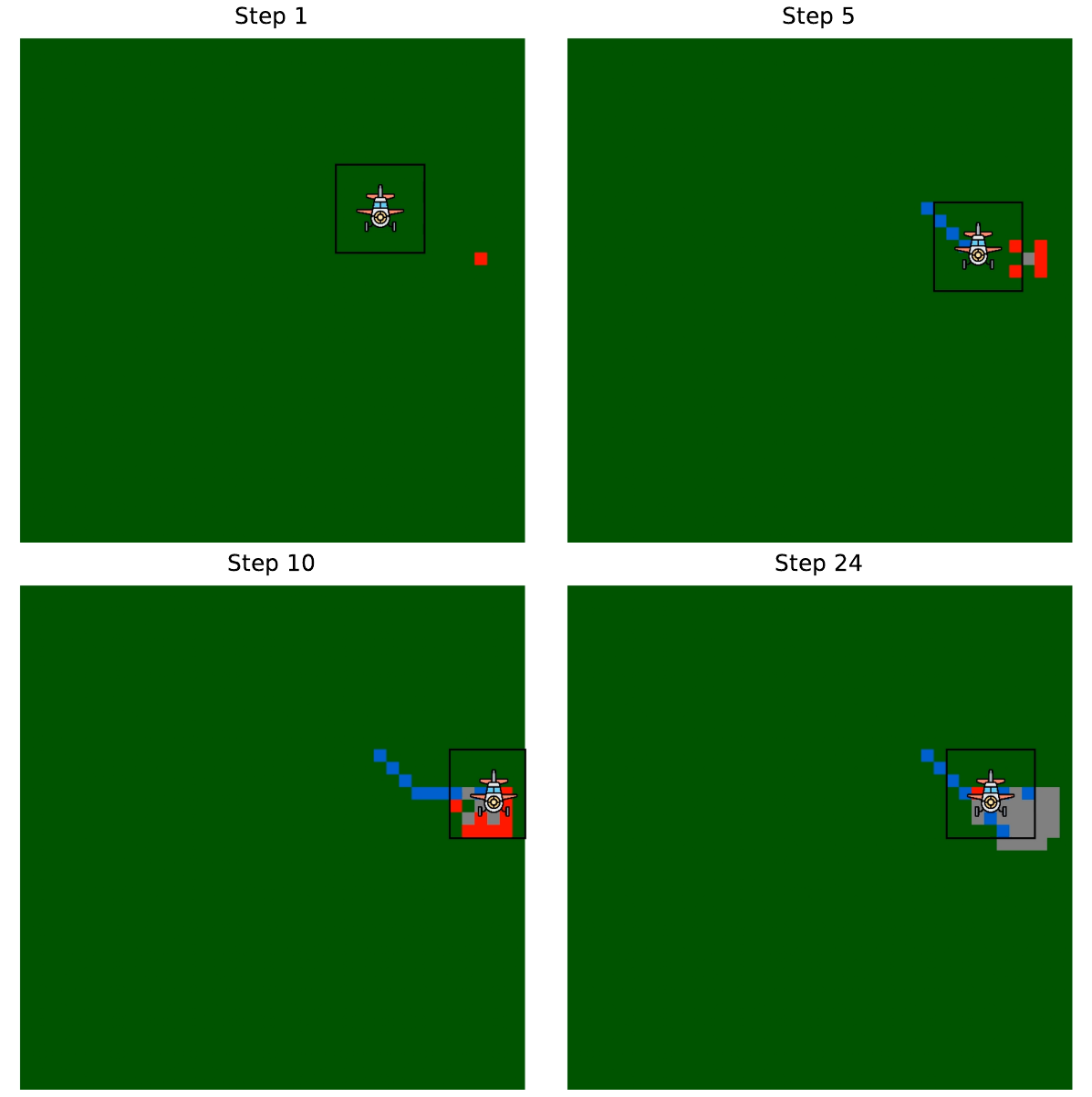}
  \caption{Illustration of the trained RL agent for fire suppression acting in the environment. It successfully learns to move towards the fire and control the valve appropriately, leading to fire extinction.}
  \label{fig:env_illustration}
\end{figure}

\end{document}